\pdfoutput=1

\documentclass[11pt]{article}
\usepackage{amsmath} 
\usepackage[htt]{hyphenat}
\usepackage{ragged2e} 
\usepackage[preprint]{acl}
\usepackage{mdframed}
\usepackage{array}
\usepackage{graphicx} 
\usepackage{float}
\usepackage{enumitem} 
\usepackage{expex}
\usepackage{svg}
\usepackage{booktabs}
\usepackage{expex}
\usepackage{xcolor}
\usepackage[T1]{fontenc}
\usepackage{listings}
\usepackage{xcolor}
\usepackage{tcolorbox}
\usepackage{multicol}
\usepackage{xcolor}       
\usepackage{xparse}       
\usepackage{svg}
\usepackage{tikz}
\usepackage{xcolor}
\usepackage{siunitx}
\usepackage{caption}
\usetikzlibrary{arrows.meta, positioning, calc, shapes.geometric}
\usepackage{xcolor}

\definecolor{acadBlue}{RGB}{0, 85, 150}
\definecolor{acadRed}{RGB}{200, 60, 60}
\definecolor{procGrey}{RGB}{245, 245, 245}
\definecolor{textGrey}{RGB}{40, 40, 40}

\usepackage{times}
\usepackage{latexsym}
\usepackage{tabularx}
\usepackage[dvipsnames]{xcolor}

\usepackage[T1]{fontenc}

\usepackage[utf8]{inputenc}

\usepackage{microtype}

\usepackage{inconsolata}

\newenvironment{xmlcode}
  {%
    \ttfamily%
  }
  {}

\newcommand{\xmltagfont}{\small}

\newcommand{\xmlopenN}[1]{%
  {\xmltagfont\textcolor{RoyalBlue}{<#1>}}%
}
\newcommand{\xmlcloseN}[1]{%
  {\xmltagfont\textcolor{RoyalBlue}{</#1>}}%
}

\newcommand{\xmlopenA}[2]{%
  {\xmltagfont
    \textcolor{RoyalBlue}{<#1}%
    \allowbreak\textcolor{BrickRed}{:#2}%
    \textcolor{RoyalBlue}{>}%
  }%
}

\newcommand{\xmlopenB}[3]{%
  {\xmltagfont
    \textcolor{RoyalBlue}{<#1}%
    \allowbreak\textcolor{BrickRed}{:#2}%
    \allowbreak\textcolor{SeaGreen}{:#3}%
    \textcolor{RoyalBlue}{>}%
  }%
}

\title{\textsc{FactAppeal}: Identifying Epistemic Factual Appeals in News Media}

\author{
    Guy Mor-Lan \qquad Tamir Sheafer \qquad Shaul R. Shenhav \\
    Hebrew University of Jerusalem \\
    \texttt{\{guy.mor, tamir.sheafer, shaul.shenhav\}@mail.huji.ac.il}
}

\begin{document}
\maketitle

\begin{abstract}
How is a factual claim made credible? We propose the novel task of \emph{Epistemic Appeal Identification}, which identifies whether and how factual statements have been anchored by external sources or evidence. To advance research on this task, we present \textsc{FactAppeal}, a manually annotated dataset of 3,226 English-language news sentences. Unlike prior resources that focus solely on claim detection and verification, \textsc{FactAppeal} identifies the nuanced epistemic structures and evidentiary basis underlying these claims and used to support them.  \textsc{FactAppeal} contains span-level annotations which identify factual statements and mentions of sources on which they rely. Moreover, the annotations include fine-grained characteristics of factual appeals such as the type of source (e.g. Active Participant, Witness, Expert, Direct Evidence), whether it is mentioned by name, mentions of the source’s role and epistemic credentials, attribution to the source via direct or indirect quotation, and other features. We model the task with a range of encoder models and generative decoder models in the 2B--9B parameter range. Our best performing model, based on Gemma 2 9B, achieves a macro-$F_1$ score of 0.73.\footnote{\href{https://github.com/guymorlan/factappeal/}{Data available at github.com/guymorlan/factappeal/}, cc-by-4.0 license.}
\end{abstract} 

\section{Introduction}

In an era marked by pervasive misinformation and heightened skepticism of media reporting, understanding how factual claims are presented has become more important than ever. While research in claim detection and verification has made substantial progress \cite{sauri2009factbank, thorne2018fever, hassan2017toward, wadden-etal-2020-fact, aly2021fact}, most existing methods focus on the content of the statements in isolation and overlook the epistemic structures that confer credibility and persuasive force to these claims. In news media, for example, the credibility of a claim is not only determined by its content but also by the way it appeals to external sources of knowledge—be it through expert testimony, official statements, or direct empirical evidence. Understanding how factual claims are anchored by appeals to external sources is also important for broader tasks in discourse analysis, fact-checking, and the study of knowledge flows in the media.

To address this gap, we introduce \textsc{FactAppeal}, a novel dataset designed to address the dual challenge of detecting both factuality and epistemic appeals within news statements. This task not only identifies whether a statement conveys a factual claim (i.e. a claim about the state of the external world) but also captures the underlying structure of how such claims are supported by sources such as experts, witnesses, and reports.

\subsection{Epistemic Appeal Identification}
An epistemic appeal is a factual claim supported by a reference to an authoritative source—whether genuinely authoritative or only purported to be—thereby providing a reason to accept the claim as true and enhancing its credibility. Epistemic appeals play a pivotal role in shaping how factual claims are constructed and perceived in public discourse, especially within news media. They are significant for analyzing epistemic justification structures for automatic fact verification, discourse analysis, and analyses of the social sources and dynamics of knowledge.

We propose the task of \textit{Epistemic Appeal Identification}, which requires determining whether a sentence presents a factual claim and, if so, identifying how it invokes an external source or evidence to support that claim. This task requires identifying the source of epistemic authority, as well as classifying the type and method of appeal. This new task pushes the boundaries of traditional factuality detection by introducing a rich layer of epistemic reasoning, crucial for understanding how information is conveyed and validated in public discourse.

\textsc{FactAppeal} comprises 3,226 sentences from news articles manually annotated with fine-grained span-level annotations. We label factual statements (whether true or false) in each sentence, as well as any epistemic appeals that ground the statement to a source of epistemic authority. We additionally include fine-grained annotations for features of epistemic factual appeals, such as the type of source invoked (e.g. Expert, Witness, Active Participant), the method of appeal (direct or indirect quotation) and more. Our annotations span a wide range of appeal types, such as official statements, reports, and testimonies, offering fine-grained insights into how claims are constructed and backed by different types of epistemic authority.

\section{Related Work}

Understanding how factual claims are supported has been the focus of several research strands in natural language processing. In this section, we review the literature on claim detection and verification, epistemic modality and argumentation, and source attribution. We then explain how our work extends these efforts by jointly modeling factuality and detailed epistemic appeals.

\subsection{Verifiability and Claim Verification}
Early work on factuality detection aimed at determining whether statements describe verifiable events \cite{sauri2009factbank, sauri-pustejovsky-2012-sure, hassan2017toward}. More recent lines of research have emphasized claim verification, exemplified by large-scale benchmarks such as FEVER \cite{thorne2018fever}, which require systems to determine if a claim is \emph{supported} or \emph{refuted} based on evidence. Other datasets have focused on specific domains or subtasks, such as SciFact \cite{wadden-etal-2020-fact} for verifying scientific claims or FactRel \cite{mor-lan-levi-2024-exploring} for factual entailment in news. While these resources have substantially advanced fact-checking methods, they focus primarily on detecting claims and modeling relations between claims, rather than providing a complete epistemic schema describing how a claim itself is constructed and supported.

\subsection{Epistemic Modality and Argumentation}

Research on epistemic modality seeks to capture linguistic markers of certainty and belief \cite{rubin2010epistemic, sauri-pustejovsky-2012-sure}, while argumentation mining explores how claims are constructed and supported within discourse \cite{feng2011argumentation}. A related task is epistemic stance detection, which models a source's degree of commitment—such as certainty or doubt—toward a claim \cite{gupta-etal-2022-examining}. Such analyses are crucial in persuasive language, where models have been trained to identify broad strategies like appeals to authority \cite{da-san-martino-etal-2019-fine-fixed}. While these works focus on belief, commitment, or general persuasive tactics, our approach offers a more granular view. By contrast, we pinpoint the \textit{concrete sources} invoked (e.g., a named expert, a witness) and classify the structural nature of the appeal itself, such as the source type and method of quotation. This allows for more precise modeling of how claims receive or signal credibility.

\subsection{Source Attribution and Quotation Analysis}
Prior studies have addressed the task of detecting quotations and attributing them to entities \cite{pareti-etal-2013-automatically, okeefe-etal-2012-sequence}, which is crucial for scientific, journalistic and legal texts. However, these methods do not typically classify sources by \emph{type} (e.g., expert vs.\ witness) or capture whether appeals are invoked through direct speech or paraphrasing. We build on these works by jointly modeling factuality and source-based epistemic appeals, thereby revealing how news articles invoke or display a source's authority to support a factual claim.

\section{Annotation Scheme}
\subsection{Overview}
We propose a span-level annotation scheme for detecting epistemic appeals in news media, labeling each relevant textual span alongside its associated features. The tags are provided both as character indices and as XML-style tags. Span-level tags are a key advantage of \textsc{FactAppeal}, allowing differentiating factual appeals, facts without appeals and non-factual components in a single text, as well as identifying multiple epistemic sources. Tags of different types may also be nested. The tags are:

\begin{itemize}
    \item \textbf{\texttt{Fact Without Appeal}} — factual claim made without epistemic appeal to a source.
    \item \textbf{\texttt{Fact With Appeal}} — factual claim made with an epistemic appeal to a source. This tag has one modifier, an additional tag for whether the identified fact reproduces the source's speech verbatim or paraphrases and processes it. It is always annotated with respect to \texttt{Fact With Appeal} spans, with two possible values:
    \begin{itemize}
        \item Direct quote
        \item Indirect quote
    \end{itemize}
    \item \textbf{\texttt{Source}} — epistemic source to which a claim is attributed. This tag has two additional modifiers annotated with respect to all identified source spans. \\ \\
    First, whether the source is mentioned by name or not:
    \begin{itemize}
        \item Named
        \item Unnamed
    \end{itemize}
    Second, the type of epistemic source:
    \begin{itemize}
        \item Active Participant
        \item Witness
        \item Direct Evidence
        \item Official
        \item Expert
        \item Expert Document
        \item News Report
        \item null (cannot be determined)
    \end{itemize}
    \item \textbf{\texttt{Source Attribute}} — marking relevant epistemic attributes of the sources, such as a title, office or status held by the epistemic source, or any information about the source cited as epistemic credentials.
    \item \textbf{\texttt{Recipient}} — recipient receiving the information from the appeal source.
    \item \textbf{\texttt{Appeal Time}} — time in which appeal was made.
    \item \textbf{\texttt{Appeal Location}} — physical, virtual or symbolic location in which appeal was made.
\end{itemize}

The primary tags are further explained below. 

\subsection{Factual Claims}

We first examine the factuality of a sentence. \textbf{Factual claims} are sentences that primarily make a statement about the state of the external world, which could be either true or false. They correspond to what Jakobson describes as the referential function of language, which is concerned with conveying information about the external world and is ``oriented toward the context'' \citep{roman1960closing}, as well as to the assertive speech act described by Searle, in which the speaker commits to the truth of what is asserted \citep{searle2013illocutionary}. 
Thus, statements that primarily convey a personal experience or subjective feeling are non-factual, and receive a null annotation:

\pex 
\texttt{\textbf{``Even so, when I visited Chennai, I felt okay about the media future we're heading into."}}
\xe

Note that the use of quotation marks does not necessitate that a cited statement is an epistemic appeal or even factual, as these categorizations depend on the dominant function of the statement.

Normative statements that primarily express a value judgment are considered non-factual within this annotation scheme:

\pex
\texttt{\textbf{They shouldn't have had anything to do with this investigation, with this case, any submission to the FISA court.}}
\xe

Similarly, questions, pleas, commands, calls to action and similar speech acts fall outside the scope of factual statements:

\pex

\a \texttt{\textbf{What exactly are you going to do?}}
\a \texttt{\textbf{Add your name to millions demanding Congress take action on the President's crimes.}}
\xe

\textbf{Factual appeals} are factual claims accompanied by a reference to a purported source of knowledge. Appeals are generally performed via some form of reference or citation,\footnote{Including unattributed quotes, in which the existence of a source is implied but its identity is not determined.} which could take the form of direct quotation reproducing speech verbatim, or indirect reference including any forms of paraphrasing or knowledge mediation.

Thus, a \textbf{brute factual statement} is a factual claim that lacks any epistemic appeal, and is annotated as follows:

\pex
\begin{xmlcode}
\xmlopenN{Fact\_No\_Appeal}
\textbf{Sometimes called street cameras, the portable P.D.Q. (Photography Done Quickly) model could produce pocket-size photographs directly onto paper, eliminating the need for negatives.}
\xmlcloseN{Fact\_No\_Appeal}
\end{xmlcode}
\xe

A challenging aspect of \textsc{FactAppeal} is distinguishing cases where an entity is mentioned merely as the subject of a report from instances where the source is cited to bolster a factual claim through its authority. For example:

\pex
\begin{xmlcode}
\xmlopenN{Fact\_No\_Appeal}
\textbf{After the successful test hop, Mr Musk said: ``One day Starship will land on the rusty sands of Mars."}
\xmlcloseN{Fact\_No\_Appeal}
\end{xmlcode}
\xe

Here, although Elon Musk is quoted, his authority is not invoked as evidence for a verifiable fact; instead, the statement primarily \textit{reports on} Musk making this comment. Consequently, this is annotated as \texttt{Fact\_No\_Appeal} rather than an epistemic appeal.

\subsection{Types of Epistemic Appeals}

We develop a structured typology of appeal sources grounded in the nature of the evidence that supports each factual claim. This framework is essential for distinguishing among various forms of authority and for clarifying how these authorities function within epistemic appeals.

As shown in Figure~\ref{fig:epistemic-appeals}, our typology classifies sources according to two fundamental dimensions: 
(1) \textit{proximity to the event (internal vs.\ external)} and 
(2) \textit{whether the source is human or non-human}. 
An internal source has a direct, firsthand connection to the event, whereas an external source provides more generalized expertise. 
\textbf{Internal appeals} thus involve a factual grounding via an epistemic source with immediate or sensory contact to the events. 
They comprise the following types:


\begin{figure*}[ht]
\centering
\resizebox{\textwidth}{!}{%

\begin{tikzpicture}[
    >=LaTeX,
    font=\sffamily\small,
    decision/.style={
        rectangle,
        draw=textGrey,
        thick,
        fill=white,
        text=textGrey,
        rounded corners=2pt,
        align=center,
        inner sep=8pt,
        text width=3cm,
        font=\sffamily\small
    },
    root/.style={
        decision,
        fill=textGrey,
        text=white,
        font=\sffamily\bfseries\small
    },
    leafInt/.style={
        rectangle,
        draw=acadRed,
        thick,
        fill=acadRed!10,
        text=acadRed,
        font=\sffamily\bfseries\small,
        rounded corners=2pt,
        align=center,
        inner sep=6pt,
        text width=2.5cm
    },
    leafExt/.style={
        rectangle,
        draw=acadBlue,
        thick,
        fill=acadBlue!10,
        text=acadBlue,
        font=\sffamily\bfseries\small,
        rounded corners=2pt,
        align=center,
        inner sep=6pt,
        text width=2.5cm
    },
    label/.style={
        font=\sffamily\scriptsize\bfseries,
        text=textGrey,
        fill=white,
        inner sep=2pt,
        opacity=1.0
    },
    line/.style={
        draw=textGrey,
        thick,
        ->,
        rounded corners=5pt
    }
]


\node [root] (source) {EPISTEMIC APPEAL\\DETECTED};

\node [decision, below=0.6cm of source] (relation) {What is the source's\\relation to the event?};

\coordinate (splitPoint) at ($(relation.south) + (0,-1.5)$);

\coordinate (leftNodePos) at ($(splitPoint) + (-5.5, -1.2)$);
\coordinate (rightNodePos) at ($(splitPoint) + (5.5, -1.2)$);

\node [decision] (intHuman) at (leftNodePos) {Is the source\\a person?};

\node [leafInt, left=1.0cm of intHuman] (direct) {Direct\\Evidence};
\node [decision, below=1.5cm of intHuman] (active) {Active involvement?};
\node [leafInt, left=1.0cm of active] (witness) {Witness};
\node [decision, below=1.5cm of active] (authority) {Has official\\authority?};

\node [leafInt, below left=1.2cm and -0.5cm of authority] (participant) {Active\\Participant};
\node [leafInt, below right=1.2cm and -0.5cm of authority] (official) {Official};

\node [decision] (extHuman) at (rightNodePos) {Is the source\\a person?};

\node [leafExt, right=1.0cm of extHuman] (expert) {Expert};
\node [decision, below=1.5cm of extHuman] (news) {Has journalistic\\authority?};

\node [leafExt, below left=1.2cm and -0.5cm of news] (expertDoc) {Expert\\Document};
\node [leafExt, below right=1.2cm and -0.5cm of news] (newsReport) {News\\Report};


\draw [line] (source) -- (relation);

\draw [line] (relation.south) -- (splitPoint) -| (intHuman.north);

\node[above, font=\sffamily\bfseries\small, text=acadRed] at ($(splitPoint)!0.5!(splitPoint-|intHuman.north)$) {INTERNAL};
\node[below, font=\sffamily\scriptsize, text=textGrey] at ($(splitPoint)!0.5!(splitPoint-|intHuman.north)$) {(Direct connection)};

\draw [line] (relation.south) -- (splitPoint) -| (extHuman.north);
\node[above, font=\sffamily\bfseries\small, text=acadBlue] at ($(splitPoint)!0.5!(splitPoint-|extHuman.north)$) {EXTERNAL};
\node[below, font=\sffamily\scriptsize, text=textGrey] at ($(splitPoint)!0.5!(splitPoint-|extHuman.north)$) {(General expertise)};

\draw [line] (intHuman) -- node[label] {NO} (direct);
\draw [line] (intHuman) -- node[label] {YES} (active);

\draw [line] (active) -- node[label] {NO} (witness);
\draw [line] (active) -- node[label] {YES} (authority);

\draw [line] (authority.south) -- ++(0,-0.5) -| node[label, pos=0.25] {NO} (participant.north);
\draw [line] (authority.south) -- ++(0,-0.5) -| node[label, pos=0.25] {YES} (official.north);

\draw [line] (extHuman) -- node[label] {YES} (expert);
\draw [line] (extHuman) -- node[label] {NO} (news);

\draw [line] (news.south) -- ++(0,-0.5) -| node[label, pos=0.25] {NO} (expertDoc.north);
\draw [line] (news.south) -- ++(0,-0.5) -| node[label, pos=0.25] {YES} (newsReport.north);

\end{tikzpicture}
}
\caption{Typology of Epistemic Appeal Sources}
\label{fig:epistemic-appeals}
\end{figure*}
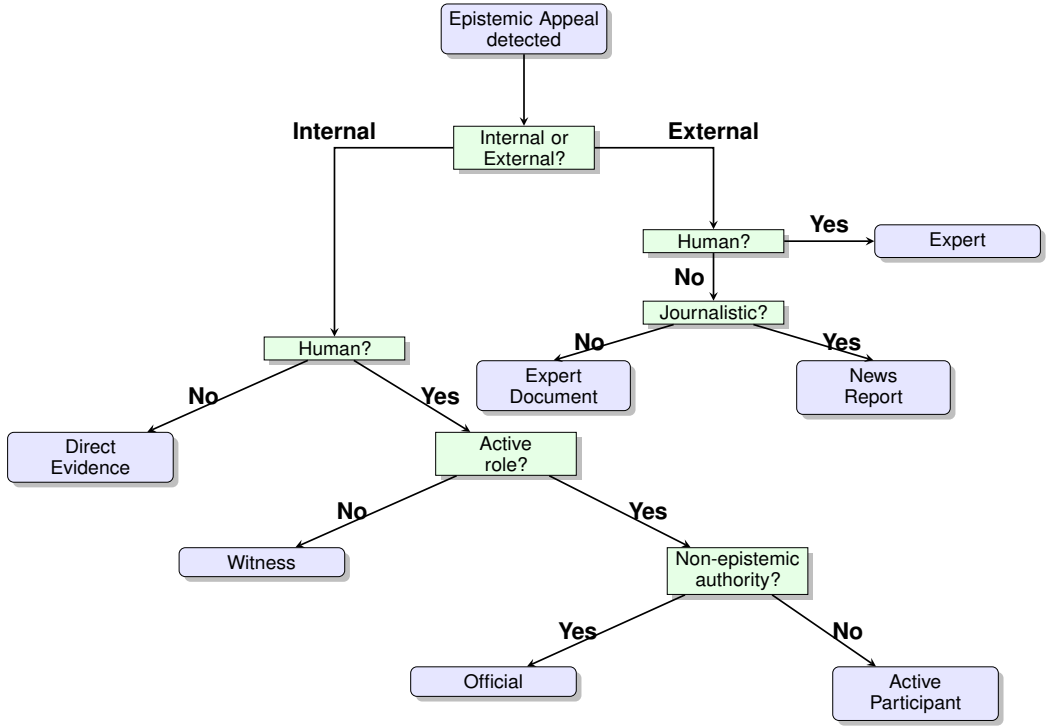
\textbf{Active participants} are actors taking active roles in the events related to the fact.

\pex
\begin{xmlcode}
\xmlopenB{Source}{Named}{Participant} \textbf{Emily} \xmlcloseN{Source} \textbf{told the} \xmlopenN{Recipient} \textbf{Buffalo News} \xmlcloseN{Recipient} \xmlopenA{Fact\_Appeal}{Indirect} \textbf{she had received a text from her mother that read: "Well, I am done with you."} \xmlcloseN{Fact\_Appeal}
\end{xmlcode}
\xe



\textbf{Witnesses} are observers who provide firsthand testimony of events but are not active participants.

\pex
\begin{xmlcode}
\textbf{Another} \xmlopenN{Source\_Attribute} \textbf{witness to the shooting,} \xmlcloseN{Source\_Attribute} \xmlopenB{Source}{Named}{Witness} \textbf{Megan Chadwick,} \xmlcloseN{Source} \textbf{said} \xmlopenA{Fact\_Appeal}{Indirect} \textbf{her husband saw the civilian take down the shooter.} \xmlcloseN{Fact\_Appeal}
\end{xmlcode}
\xe

\textbf{Officials} are active participants who also have extra non-epistemic authority on events or on facts — e.g., legal, political, bureaucratic authority. Officials, such as government authorities, often provide statements that carry legal or formal weight. Importantly, officials wield power that can alter states of affairs related to the factual claim.

\pex
\begin{xmlcode}
\xmlopenB{Source}{Named}{Official} \textbf{Doug Ericksen,} \xmlcloseN{Source} \xmlopenN{Source\_Attribute} \textbf{the EPA's communications director for the transition,} \xmlcloseN{Source\_Attribute} \textbf{told} \xmlopenN{Recipient} \textbf{National Public Radio} \xmlcloseN{Recipient} \textbf{that} \xmlopenA{Fact\_Appeal}{Direct} \textbf{``we'll take a look at what's happening so that the voice coming from the EPA is one that's going to reflect the new administration."} \xmlcloseN{Fact\_Appeal}
\end{xmlcode}
\xe

\textbf{Direct evidence} is an appeal to a piece of evidence found ``at the scene'' and bearing on the facts related to the factual claim.

\pex
\begin{xmlcode}
\xmlopenB{Source}{Unnamed}{Direct\_Evidence} \textbf{This 2013 photo} \xmlcloseN{Appeal\_Source} \textbf{provided to} \xmlopenN{Recipient} \textbf{The Associated Press} \xmlcloseN{Recipient} \textbf{shows} \xmlopenA{Fact\_Appeal}{Indirect} \textbf{now-defrocked Catholic priest Richard Daschbach leading a service at a church in Kutet, East Timor.} \xmlcloseN{Fact\_Appeal}
\end{xmlcode}
\xe

\textbf{External appeals} on the other hand involve appeals to a source without a firsthand connection to events, whose epistemic credentials are grounded in general expertise. These sources possess epistemic expertise which bears on the factual claim:

\textbf{Experts}, such as scientists or specialists, offer appeals rooted in professional expertise and specialized knowledge.

\pex
\begin{xmlcode}
\xmlopenA{Fact\_Appeal}{Direct} \textbf{``The dolphins of Sarasota Bay are really good indicators of the health of our ecosystem,"} \xmlcloseN{Fact\_Appeal} \textbf{said} \xmlopenB{Source}{Named}{Expert} \textbf{Dr. Wells.} \xmlcloseN{Source}
\end{xmlcode}
\xe

\textbf{Expert Document} refers to expert knowledge embodied in non-human objects, such as research documents, scientific and institutional reports.

\pex
\begin{xmlcode}
\textbf{A 2013} \xmlopenB{Source}{Unnamed}{Expert\_Doc} \textbf{study} \xmlcloseN{Source} \textbf{found that} \xmlopenA{Fact\_Appeal}{Indirect} \textbf{peppermint oil has potent antiseptic properties which are useful against oral pathogens.} \xmlcloseN{Fact\_Appeal}
\end{xmlcode}
\xe

\textbf{News Report} refers to citation of previous news reports.

\pex
\begin{xmlcode}
\xmlopenA{Fact\_Appeal}{Direct} \textbf{"We are praying to God and asking that all Argentines help us to pray that they keep navigating and that they can be found,"} \xmlcloseN{Fact\_Appeal} \xmlopenB{Appeal\_Source}{Named}{Witness} \textbf{Claudio Rodriguez} ,\xmlcloseN{Appeal\_Source} \xmlopenN{Source\_Attribute} \textbf{the brother of one of the crew members} \xmlcloseN{Source\_Attribute} \textbf{told} \xmlopenN{Recipient} \textbf{the local Todo Noticias TV channel,} \xmlcloseN{Recipient} \textbf{according to} \xmlopenB{Appeal\_Source}{Named}{News\_Report} \textbf{the AP.} \xmlcloseN{Appeal\_Source}
\end{xmlcode}
\xe

The distinction between internal and external sources also reflects two modes or logics of epistemology —a common-wisdom logic preferring those with direct relations to the matter at hand, as opposed to an expertise-based logic preferring ``detached" experts. Whereas internal sources have epistemic credentials in virtue of their specific history and contact with the situation at hand, external sources possess epistemic credibility due to their attained expertise \citep{pierson1994epistemic, collins2002third}.

\section{Dataset}

The dataset contains 3,226 sentences sampled from diverse English-language news articles published between 2020 and 2022. Each sentence was annotated by one of two annotators: one of the authors and a student research assistant (see appendices \ref{sec:appendix-annotators} and \ref{sec:annotation-guidelines}). The dataset has been randomly split into a training set (70\%), development set (15\%) and test set (15\%).

\subsection{Inter-Annotator Agreement Analysis}

We conducted an inter-annotator agreement (IAA) analysis on a subset of data labeled by both annotators. To facilitate the comparison, each span annotation was converted into binary word-level labels. Using these labels, we computed several metrics—namely the union and intersection counts, the number of words where neither annotator marked the tag, the Intersection over Union (IoU), and Cohen's Kappa. Table~\ref{tab:agreement_metrics} summarizes the IAA statistics for each tag. The overall IoU of 0.74 and a Cohen's Kappa of 0.82 indicate substantial agreement between the annotators. However, some span annotations are relatively rare and have few instances.



\begin{table}[ht]
\centering
\resizebox{\columnwidth}{!}{%
\begin{tabular}{
  l
  S[table-format=4.0]
  S[table-format=4.0]
  S[table-format=4.0]
  S[table-format=1.2]
  S[table-format=1.2]
}
\toprule
Tag & {Union} & {Intersection} & {Unlabeled} & {IoU} & {Cohen's $\boldsymbol{\kappa}$} \\
\midrule
Fact w/o Appeal   & 511  & 372  & 1194 & 0.73 & 0.79 \\
Fact with Appeal  & 986  & 732  & 719  & 0.74 & 0.70 \\
Appeal Time       & 15   & 11   & 1690 & 0.73 & 0.85 \\
Appeal Location   & 27   & 17   & 1678 & 0.63 & 0.77 \\
Recipient         & 14   & 14   & 1691 & 1.00 & 1.00 \\
Source            & 131  & 104  & 1574 & 0.79 & 0.88 \\
Source Attribute  & 90   & 83   & 1615 & 0.92 & 0.96 \\
Indirect Quote    & 669  & 420  & 1036 & 0.63 & 0.67 \\
Direct Quote      & 368  & 261  & 1337 & 0.71 & 0.79 \\
Active Participant& 21   & 12   & 1684 & 0.57 & 0.73 \\
Witness           & 22   & 19   & 1683 & 0.86 & 0.93 \\
Direct Evidence   & 20   & 13   & 1685 & 0.65 & 0.79 \\
Official          & 23   & 15   & 1682 & 0.65 & 0.79 \\
Expert            & 14   & 12   & 1691 & 0.86 & 0.92 \\
External Document & 27   & 15   & 1678 & 0.56 & 0.71 \\
\bottomrule
\end{tabular}%
}
\caption{Word-level Inter-annotator Agreement Metrics\protect\footnotemark}
\label{tab:agreement_metrics}
\end{table}

\footnotetext{\texttt{Named}/\texttt{Unnamed} are excluded as they were added later. \texttt{External Document} includes both \texttt{Expert Document} and \texttt{News Report}, which were later split from this category.}

\subsection{Descriptive Statistics}

\begin{figure}[htbp]
    \centering
    \includegraphics[width=0.5\textwidth]{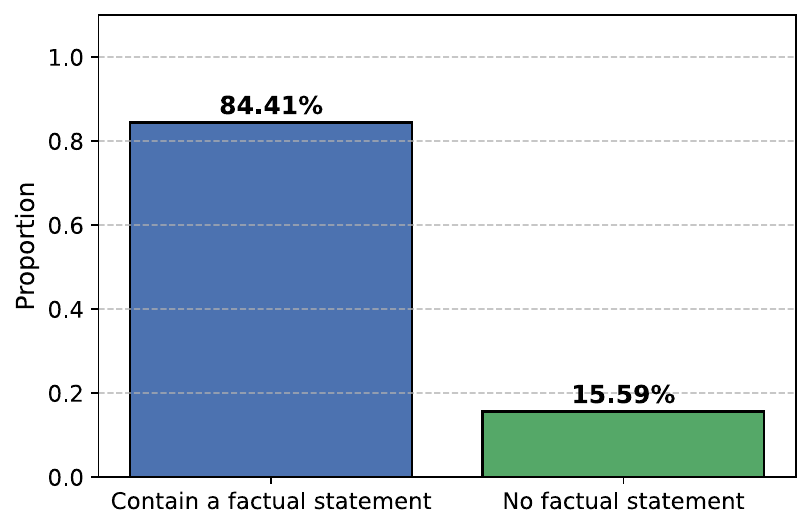}
    \caption{Proportion of Factual Sentences}
    \label{fig:distribution_plots}
\end{figure}

We examine the share of sentences containing any factual claim in Figure~\ref{fig:distribution_plots}. More than 80\% of statements are annotated as factual. While this may seem high, it corresponds well to the factual transmitting nature of news reports.

\begin{figure*}[!h]
    \centering
\includegraphics[width=0.85\textwidth]{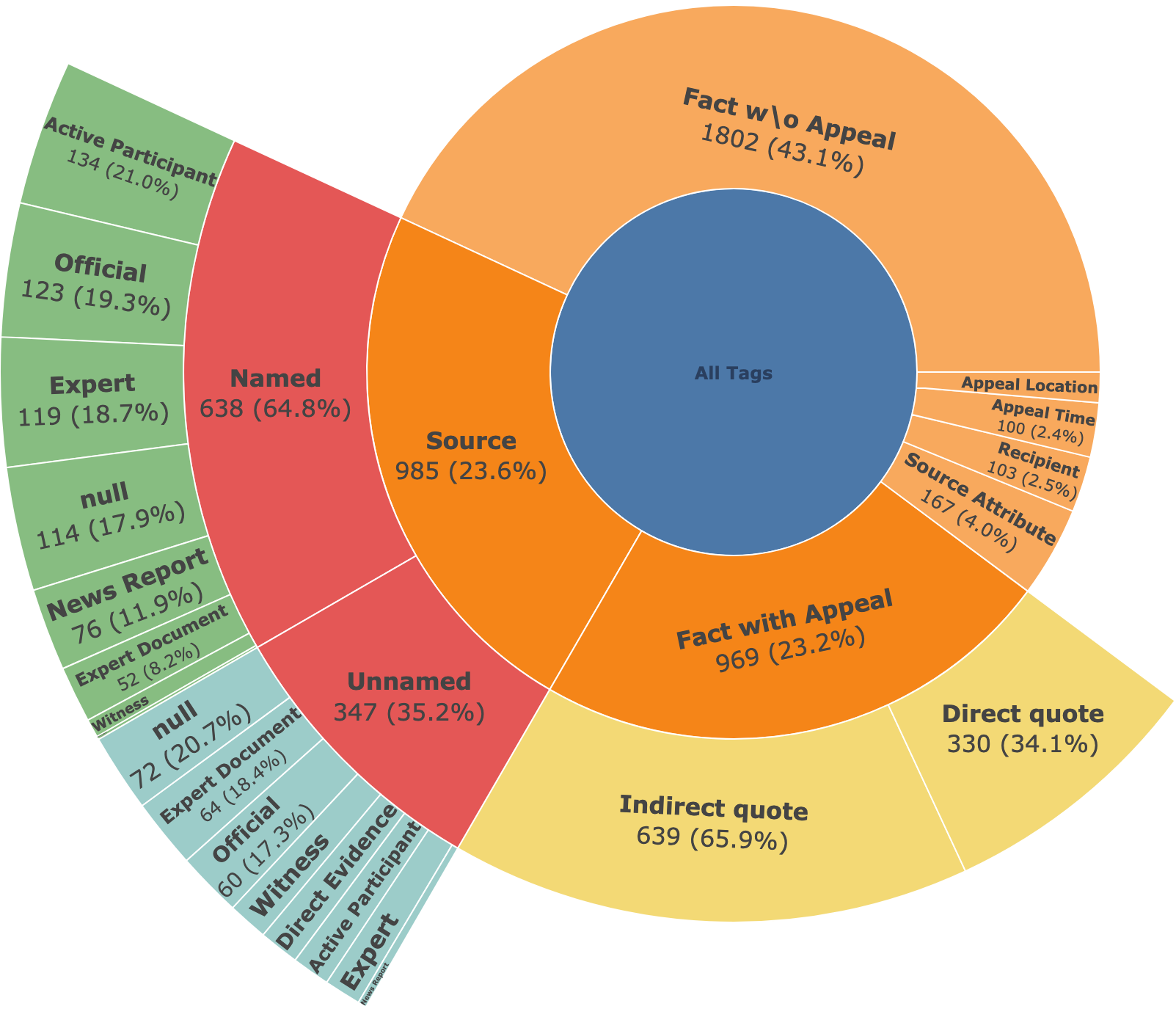}
    \caption{Distribution of Span Tags\protect\footnotemark}
    \label{fig:tag_distribution}
\end{figure*}

In Figure~\ref{fig:tag_distribution}, we present the distribution of span annotations. We first observe that statements without epistemic appeals appear nearly twice as frequently as those containing appeals. Moreover, we observe that most factual appeals utilize paraphrasing (66\%) rather than direct quotation (34\%).

When an appeal source is mentioned, it is usually mentioned by name (64\%). For named sources, the most popular types are active participants (20\%), news reports and expert documents (20\%), officials (19\%) and experts (19\%), and are thus almost equally prevalent. Witnesses and direct evidence account for a smaller share and thus appear substantially less common as sources of knowledge.

For appeal sources that are unnamed (35\%), news reports and expert documents are most common (24\%), followed by the null category for indeterminate types (19\%) and officials (17\%). Witnesses, experts, active participants and direct evidence thus appear less frequently as unnamed sources.


\section{Experiments}

We compare two modeling strategies for Epistemic Appeal Identification (see Appendix \ref{sec:setup}):

\paragraph{Token-level multi-label classification with encoder models.} Since different tag types may overlap, in this setting we represent the tags as token-level multi-label binary annotations, with 18 labels corresponding to each of the tags and possible modifier values. We fine-tune pre-trained transformer encoder models, using the base model versions of RoBERTa \citep{liu2019robertarobustlyoptimizedbert}, DeBERTa v3 \citep{he2021debertadecodingenhancedbertdisentangled} and ModernBERT \citep{warner2024smarter}. The encoder models are trained for up to 12 epochs with focal loss \citep{lin2017focal}.
    
\paragraph{Generative decoder models.} In this setting, annotations are represented as XML-style tags (similar to the presentation in Section 3). Models are trained to produce the annotated sentence given the raw sentence. We fine-tune several smaller pre-trained LLMs such as Gemma 2 (2B and 9B) \citep{team2024gemma}, Llama 3.1 8B \citep{grattafiori2024llama} and Mistral v0.3 7B \citep{jiang2023mistral7b}. The models are trained with QLORA \citep{dettmers2023qloraefficientfinetuningquantized} with 4-bit quantization for 3 epochs, with $r=256$ and $\alpha=256$. We mask the loss of the input prompt and train on completions.

In order to allow for partial matches and robustness across tokenizers, we calculate word-level scores for each of the 18 tag categories. Table \ref{tab:global_metrics} reports macro-average word-level precision, recall and $F_1$ scores on the test set. While differences between top encoder models and top decoder models are small, the largest decoder, Gemma 2 9B, achieves the best macro-$F_1$ score of 0.73.

In Table \ref{tab:per_tag_f1} we take a look at the $F_1$ scores per tag in the test set. While some tags are learned well by encoder models, encoder models show higher variation in performance across categories. Moreover, encoder models show a stronger correlation between tag counts and test $F_1$ scores ($\rho_s = 0.72$) than decoders ($\rho_s = 0.66$).

For source type annotations specifically, performance is less correlated with the prevalence of the tags, as the four more prevalent tags \textit{Active Participant}, \textit{Expert Document}, \textit{Official} and \textit{Expert} are not necessarily better detected than the less prevalent \textit{Direct Evidence} and \textit{Witness}. Here again, tag prevalence is more strongly correlated with $F_1$ scores for encoder models ($\rho_s = 0.18$) and less so for decoders ($\rho_s = 0.07$).

Overall, these results indicate that \textit{Epistemic Appeal Identification} remains challenging for encoder-only models and smaller LLMs, highlighting significant room for improvement.

\begin{table}[ht]
\centering
\resizebox{\columnwidth}{!}{%
\begin{tabular}{lccc}
\toprule
Model & Precision & Recall & $F_1$ \\
\midrule
RoBERTa (base)     & 0.75 & 0.67 & \textbf{0.7} \\
DeBERTa v3 (base)  & 0.73 & 0.67 & 0.69 \\
ModernBERT (base)  & 0.73 & 0.47 & 0.54 \\

\midrule

Gemma 2 9B         & 0.76 & 0.73 & \textbf{0.73} \\
Mistral v0.3 7B    & 0.73 & 0.68 & 0.7 \\
Llama 3.1 8B       & 0.75 & 0.65 & 0.68 \\
Gemma 2 2B         & 0.65 & 0.58 & 0.6 \\

\bottomrule
\end{tabular}%
}
\caption{Global Macro Metrics, Test Set}
\label{tab:global_metrics}
\end{table}

\bigskip

\begin{table*}[htbp]
\centering
\small 
\resizebox{\textwidth}{!}{%
\sisetup{
  table-number-alignment = center,
  round-mode           = places,
  round-precision      = 2,
  table-space-text-post = {}, 
}
\begin{tabular}{
  l
  S[table-format=1.2]
  S[table-format=1.2]
  S[table-format=1.2]
  S[table-format=1.2]
  S[table-format=1.2]
  S[table-format=1.2]
  S[table-format=1.2]
}
\toprule
Tag & {RoBERTa} & {ModernBERT} & {DeBERTa v3} & {Gemma 2 2B} & {Llama 3.1 8B} & {Mistral v0.3 7B} & {Gemma 2 9B} \\
\midrule
\multicolumn{8}{l}{\textbf{Factuality}} \\
Fact w/o Appeal  & 0.87 & 0.83 & 0.86 & 0.84 & 0.86 & 0.88 & \textbf{0.89} \\
Fact with Appeal & 0.84 & 0.84 & \textbf{0.86} & 0.78 & 0.83 & 0.84 & 0.85 \\
\addlinespace
\multicolumn{8}{l}{\textbf{Appeal Characteristics}} \\
Appeal Time      & 0.75 & 0.69 & 0.65 & 0.44 & 0.60 & 0.68 & \textbf{0.77} \\
Appeal Location  & 0.47 & 0.43 & 0.49 & 0.31 & 0.57 & 0.47 & \textbf{0.58} \\
Recipient        & 0.77 & 0.51 & \textbf{0.90} & 0.62 & 0.71 & 0.81 & 0.89 \\
Source           & 0.81 & 0.77 & \textbf{0.84} & 0.70 & 0.77 & 0.79 & \textbf{0.84} \\
Source Attribute & 0.71 & 0.53 & \textbf{0.80} & 0.65 & 0.66 & 0.58 & 0.79 \\
\addlinespace
\multicolumn{8}{l}{\textbf{Quotation Type}} \\
Indirect Quote   & 0.83 & 0.83 & \textbf{0.87} & 0.77 & 0.82 & 0.81 & 0.83 \\
Direct Quote     & 0.77 & 0.79 & 0.79 & 0.72 & 0.76 & 0.78 & \textbf{0.80} \\
\addlinespace
\multicolumn{8}{l}{\textbf{Source Named}} \\
Named            & 0.75 & 0.76 & \textbf{0.80} & 0.68 & 0.71 & 0.72 & 0.77 \\
Unnamed          & \textbf{0.69} & 0.53 & 0.67 & 0.49 & \textbf{0.69} & 0.64 & 0.65 \\
\addlinespace
\multicolumn{8}{l}{\textbf{Source Type}} \\
Active Participant & 0.43 & 0.16 & 0.26 & 0.40 & 0.45 & 0.44 & \textbf{0.54} \\
Witness            & 0.66 & 0.10 & 0.55 & 0.57 & 0.57 & \textbf{0.81} & 0.57 \\
Direct Evidence    & 0.58 & 0.19 & 0.51 & 0.30 & \textbf{0.73} & 0.57 & 0.71 \\
Official           & 0.67 & 0.51 & 0.62 & 0.59 & 0.59 & 0.59 & \textbf{0.71} \\
Expert             & 0.60 & 0.17 & 0.57 & 0.52 & \textbf{0.65} & 0.63 & 0.62 \\
Expert Document  & 0.61 & 0.43 & 0.62 & 0.63 & 0.67 & \textbf{0.73} & 0.68 \\
News Report        & 0.76 & 0.68 & 0.75 & \textbf{0.79} & 0.68 & 0.78 & 0.68 \\
\midrule
\multicolumn{1}{l}{\textit{Standard Deviation}} & 0.12 & 0.25 & 0.17 & 0.16 & 0.10 & 0.13 & 0.11 \\
\bottomrule
\end{tabular}
}
\caption{Per-Tag $F_1$ Scores}
\label{tab:per_tag_f1}
\end{table*}

\section{Conclusion} 
In this work, we introduce \textsc{FactAppeal}, a novel dataset and task formulation aimed at identifying epistemic appeals in news media factual claims. Our dataset captures both the factuality of claims and the underlying epistemic structures that lend these claims credibility. The experiments comparing token-level predictions using encoder models with generative LLMs underscore the challenges of modeling nuanced epistemic appeals, as well as the strength of generative models and the feasibility of sequence-to-sequence representations for this task.

Beyond advancing the modeling of epistemic appeals, this work also contributes to the fields of factual detection and automated fact-checking, offering span-level annotations that capture how claims are justified in news media. By providing fine-grained annotations—differentiating factual from non-factual statements and detailing the types of epistemic appeals—our approach opens new avenues for more context-aware fact-checking. This dual focus on both factuality and the structure of supporting evidence addresses key limitations in current factuality detection frameworks and paves the way for more robust news factuality analysis.

Furthermore, \textsc{FactAppeal} has important implications for social science research across political philosophy, social epistemology, and communication. Scholars such as Anderson \cite{Anderson2021-ANDEBA-3} and Lynch \cite{Lynch2021-LYNPDA} have highlighted that contrasting epistemic frameworks can lead to ``deep disagreements'' among political groups, and communication scholars have underscored the central role of media in shaping which facts gain prominence and how audiences interpret them \cite{mccombs1972agenda, entman1993framing}. More recent studies demonstrate how the information environment influences factual beliefs, partisan divides, and public polarization \cite{jerit2012partisan, aalberg2012media, garrett2016driving, djerf2017still}. By systematically identifying and modeling epistemic appeals, \textsc{FactAppeal} offers a powerful tool for investigating how news media construct and validate factual claims—a process fundamental to understanding broader social dynamics and shifts in political discourse.

Future research can leverage these contributions in several ways. In factuality and fact-checking, our dataset may improve claim verification and evidence detection approaches by incorporating source-based credibility cues. Extending \textsc{FactAppeal} to larger textual units, such as paragraphs or entire articles, could reveal more complex discourse structures and further enhance automated verification. In computational discourse analysis, \textsc{FactAppeal} can facilitate deeper investigations of epistemic appeals in public discourse, shedding light on broader patterns of justification, knowledge transfer, and media polarization.

Secondly, the community can utilize \textsc{FactAppeal} to refine factual appeal modeling even further—by exploring appeals in larger contexts, linking multiple sources and claims, and identifying additional attributes of factual epistemic appeals. Future expansions could also include social media content and other distinct types of discourse.

\clearpage

\section*{Limitations}
While \textsc{FactAppeal} marks an important step forward in capturing epistemic structures in news media, our work has several limitations. First, the dataset employs only sentence-level annotations, which restricts the amount of contextual information that can be captured. Future studies might extend annotations to paragraphs or entire articles, where relationships among claims, sources, and evidence can be modeled more comprehensively. 

Second, although multiple sources or factual claims can appear in a single sentence, the current annotations do not explicitly link each source to its corresponding claim. Such explicit linkage could improve the granularity of epistemic appeal analyses and enable more precise modeling of how diverse sources relate to one or more claims within the same sentence.

Finally, \textsc{FactAppeal} comprises English-language news articles from a particular time frame (2020--2022). This narrow focus may limit the generalizability of our findings to other languages, domains, or historical periods. Future research could address these limitations by applying the annotation scheme to broader contexts and by leveraging multilingual corpora. 

\section*{Acknowledgments}

This work was supported by the Israel Science Foundation (Grant no. 2501/22). We thank Hagar Kaminer for diligent research assistance, the members of the DeepStory lab, and especially Effi Levi for insightful comments and suggestions that improved the final version of this manuscript.

\bibliography{anthology,custom}

\appendix

\section{Annotators}
\label{sec:appendix-annotators}

The dataset has been annotated by two annotators, one of the authors and a paid student research assistant. The annotators are a man and a woman in their 20s--30s from the EMEA region.

\section{Annotation Guidelines}
\label{sec:annotation-guidelines}

These guidelines describe what constitutes a factual statement, how to detect whether it appeals to an external source, and how to label the source and its attributes.  They also detail how to mark the relevant spans in the text.

\subsection{Determining Factuality}
\paragraph{Definition.} A sentence is \textbf{factual} if it primarily makes a statement about the external world that can be objectively true or false. 
Statements focusing on subjective feelings, judgments, calls to action, or questions generally do \textbf{not} count as factual for this annotation scheme.

\paragraph{Label.} 
\begin{itemize}[label=--, nosep, leftmargin=1em]
\item \texttt{Fact\_No\_Appeal} (``Fact Without Appeal'') for factual statements that do not cite an external source.
\item \texttt{Fact\_Appeal} (``Fact With Appeal'') for factual statements that explicitly reference an external source or evidence to support their claim.
\end{itemize}

\paragraph{Non-Factual Content.} If a sentence is \emph{primarily} non-factual (for instance, it is dominated by a personal opinion or call to action), it receives no fact-related annotation. 

\subsection{Identifying Epistemic Appeals}
\paragraph{Definition.} An \textbf{epistemic appeal} is a factual claim that is accompanied by a reference to an external source or evidence. The reference can be direct (quoted verbatim) or indirect (paraphrased or summarized). 

\paragraph{Distinguishing Reporting from Appeals.} When a statement merely \emph{covers} someone's words or remarks without using the speaker’s position or information as evidence for a factual claim, it is annotated as \texttt{Fact\_No\_Appeal}. By contrast, if the statement explicitly \emph{invokes} external authority or specialized knowledge as the reason to accept the factual claim, it is \texttt{Fact\_Appeal}. 

\paragraph{Method of Appeal.} For each \texttt{Fact\_Appeal} span, annotate the manner in which the claim references its source:
\begin{itemize}[label=--, nosep, leftmargin=1em]
    \item \texttt{Direct} (quoted verbatim)
    \item \texttt{Indirect} (paraphrased or mediated)
\end{itemize}

\subsection{Source Annotations}
\label{sec:source-types}

When annotating a \texttt{Fact\_Appeal} span, identify the \texttt{Source} span(s) explicitly referenced in that statement. The \texttt{Source} tag has two modifiers:

\paragraph{Source Name.} 
\begin{itemize}[label=--, nosep, leftmargin=1em]
    \item \texttt{Named}: The text gives a proper name or explicit identity.
    \item \texttt{Unnamed}: The source is referenced, but not by name (e.g., ``an official stated...'').
\end{itemize}

\paragraph{Source Type.} Each source is labeled with one of the following:
\begin{itemize}[label=--, nosep, leftmargin=1em]
    \item \texttt{Active\_Participant}: Has a direct, primary role in the events in question. 
    \item \texttt{Witness}: Observed the events but was not directly involved. 
    \item \texttt{Direct\_Evidence}: A non-human piece of evidence (e.g., footage, photograph) closely tied to the scene. 
    \item \texttt{Official}: Holds a position of non-epistemic authority (legal, governmental, etc.). 
    \item \texttt{Expert}: A person with specialized knowledge not derived from direct involvement (e.g., scientist, analyst). 
    \item \texttt{Expert\_Document}: A written or recorded source of expertise (e.g., a published paper). 
    \item \texttt{News\_Report}: A journalistic source. 
    \item \texttt{null}: Source type cannot be determined.
\end{itemize}

\subsection{Additional Attributes}
If relevant information is present, you may also label the following:

\begin{itemize}[leftmargin=2em]
    \item \textbf{\texttt{Source\_Attribute}}: Any text specifying the authority, rank, credentials, or role of the source (e.g., an official title).
    \item \textbf{\texttt{Recipient}}: The entity or individual to whom the source directed the claim (if explicitly stated).
    \item \textbf{\texttt{Appeal\_Time}}: The time when the appeal was made (if explicitly mentioned, e.g., “yesterday” or a date).
    \item \textbf{\texttt{Appeal\_Location}}: The physical, virtual, or symbolic location (e.g., “during a press briefing at the White House”).
\end{itemize}

\subsection{Marking the Spans}
All annotations should be represented at the span level. Spans can overlap or nest. For instance, a \texttt{Fact\_Appeal} span could contain one or more \texttt{Source} sub-spans. Make sure each factual statement is fully wrapped, and all relevant sources or attributes within it are separately tagged.

\subsection{Edge Cases and Practical Tips}
\paragraph{Multiple Sources or Claims.}
A single sentence may present more than one claim or more than one source. Tag each factual claim with or without appeal separately. If a sentence has multiple appeals or different source types, annotate each source individually.

\paragraph{Attribution Without Clear Source Type.}
If the text provides insufficient detail to determine the source type (e.g., just “sources say...” with no additional context), use the \texttt{null} label for \texttt{Source\_Type}.

\paragraph{Unclear Factuality or Mixed Content.}
If the sentence intermixes factual and non-factual statements, identify which portion is factual, provided it constitutes a coherent factual claim. Non-factual segments do not receive tags.

\section{Experimental Setup}
\label{sec:setup}

Models were trained on a single A100 GPU with 40GB VRAM, with the longest model run taking 4 hours to complete. A learning rate of 1e-5 was used. The results in the paper correspond to a single run.

For decoder models, we used a batch size of 6. The AdamW optimizer was used with the default Huggingface settings.

We used the following prompt template for training and inference:
\{input\_sentence\}\textbackslash n\#\#\# Answer: \{annotated\_sentence\}

The input sentence tokens were masked from loss calculation.

Table \ref{tab:model_sizes} documents the number of parameters in the models utilized in the experiments.

\begin{table}[ht]
\centering
\small
\begin{tabular}{l c}
\toprule
\textbf{Model} & \textbf{Parameter Size} \\
\midrule
RoBERTa (base) & 125M \\
ModernBERT (base) & 150M \\
DeBERTa v3 (base) & 184M \\
Gemma 2 2B & 2.2B \\
Mistral v0.3 7B & 7.0B \\
Llama 3.1 8B & 8.0B \\
Gemma 2 9B & 9.0B \\
\bottomrule
\end{tabular}
\caption{Parameter sizes of models used in experiments.}
\label{tab:model_sizes}
\end{table}

\end{document}